\documentclass{article}

\usepackage[preprint]{neurips_2023}

\usepackage[utf8]{inputenc} % allow utf-8 input
\usepackage[T1]{fontenc}    % use 8-bit T1 fonts
\usepackage{hyperref}       % hyperlinks
\usepackage{url}            % simple URL typesetting
\usepackage{booktabs}       % professional-quality tables
\usepackage{amsfonts}       % blackboard math symbols
\usepackage{nicefrac}       % compact symbols for 1/2, etc.
\usepackage{microtype}      % microtypography
\usepackage{xcolor}         % colors

\usepackage{graphicx}
\usepackage{multirow}
\usepackage{amssymb}
\usepackage{adjustbox}
\usepackage{colortbl}
\usepackage{amsmath}
\usepackage{enumitem}
\setlist{nolistsep}

\title{MedBLIP: Bootstrapping Language-Image Pre-training from 3D Medical Images and Texts}

\author{%
 Qiuhui Chen, Xinyue Hu, Zirui Wang, Yi Hong\thanks{Corresponding Author: yi.hong@sjtu.edu.cn} \\
  Shanghai Jiao Tong University\\
}

\begin{document}

\maketitle

\begin{abstract}
Vision-language pre-training (VLP) models have been demonstrated to be effective in many computer vision applications. In this paper, we consider developing a VLP model in the medical domain for making computer-aided diagnoses (CAD) based on image scans and text descriptions in electronic health records, as done in practice. To achieve our goal, we present a lightweight CAD system MedBLIP, a new paradigm for bootstrapping VLP from off-the-shelf frozen pre-trained image encoders and frozen large language models. We design a MedQFormer module to bridge the gap between 3D medical images and 2D pre-trained image encoders and language models as well. To evaluate the effectiveness of our MedBLIP, we collect more than 30,000 image volumes from five public Alzheimer's disease (AD) datasets, i.e., ADNI, NACC, OASIS, AIBL, and MIRIAD. On this largest AD dataset we know, our model achieves the SOTA performance on the zero-shot classification of healthy, mild cognitive impairment (MCI), and AD subjects, and shows its capability of making medical visual question answering (VQA). The code and pre-trained models is available online: \href{https://github.com/Qybc/MedBLIP}{https://github.com/Qybc/MedBLIP}. 

%to generalize the concept of BLIP-2~\cite{li2023blip} to 3D medical image domain. Inspired by BLIP-2, we apply Learnable Queries to interact with 3D medical image and corresponding clinical reports respectively, to generate the most useful 3d representations. We conduct thorough experiments to validate the effectiveness of our proposed architecture, and benchmark on numerous public benchmarks e.g., ADNI, NACC, OASIS, AIBL, and MIRIAD. Extensive experiments demonstrate that the proposed MedBLIP pretraining strategy significantly improves the performance of 3D medical diagnosis. The code and pre-trained models are available online at ***. % https://github.com/Qybc/MedBLIP.
\end{abstract}

\section{Introduction}
Electronic health records (EHR), e.g., radiology images, lab and test results, and patient demographics, are often used in clinical diagnosis. For instance, to diagnose Alzheimer's Disease (AD), apart from brain imaging, physicians also use physical and neurological exams and diagnostic tests, with these test results presented in the text form. In past decades, researchers have gradually collected a large number of EHRs, e.g., ADNI~\cite{petersen2010alzheimer}, NACC~\cite{beekly2007national}, OASIS~\cite{marcus2007open}, for studying AD. However, learning how to make diagnoses based on these EHRs, especially how to fuse these medical data from different resources and in different forms, e.g., images and texts, is still a challenging task in computer-aided diagnosis (CAD). 

Recently, large vision language pre-training (VLP) models, e.g., CLIP~\cite{radford2021learning}, BLIP~\cite{li2022blip}, BLIP-2~\cite{li2023blip}, have achieved great success in many downstream computer vision applications, such as classification~\cite{bao2022vlmo}, segmentation~\cite{xu2021simple}. These VLP models learn multi-modal representations from large image and text datasets, by aligning their features into a common space for learning. In the medical domain, researchers propose Medical Bootstrapping Language-Image Pre-training (MedCLIP)~\cite{wang2022medclip}, which learns generic representation from large-scale medical image-text pairs. This pre-trained medical model presents its generalization to various medical tasks, especially where limited medical data or labels are available for learning. However, most existing VLP models handle the situation that texts are corresponding textual descriptions of their paired images, such as image captions or medical reports. 

%In this scenario, texts have a weak connection with their paired medical images, which challenges current VLP models since they may struggle to accurately describe an image or generate meaningful text.  

\begin{figure*}[t]
  \centering
  \includegraphics[width=\linewidth]{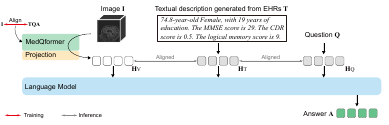}
  \caption{Architecture overview of our proposed MedBLIP, a CAD system designed for medical diagnosis with electronic health records via multimodel representation learning in a language model.}
  \label{fig:cad}
\end{figure*}

%besides textual descriptions that reflect the content of structures in medical scans,
In this paper, we consider another scenario where images and texts provide complementary information, that is, texts include additional information except for medical scans in EHRs, e.g., the age, gender, and lab results of a subject, to make an informed CAD decision. Our goal is to learn a VLM that suits this CAD scenario, which has multi-model intelligence to fuse different types of medical data, e.g., 3D medical scans and texts that contain complementary information from EHRs for CAD. Here, we need to address three problems: (1) How to extend a 2D image encoder to extract features from 3D medical images? (2) How to align image and text features and learn multi-model representations? (3) How to obtain a lightweight language model for our CAD purpose? Inspired by BLIP-2~\cite{li2023blip}, we propose MedBLIP as shown in Fig.~\ref{fig:cad}, a bootstrapping language-image pre-training model to fuse 3D medical images and texts based on a query mechanism. We first adopt a learnable patch embedding to bridge the gap between 3D medical images and a pre-trained image encoder, which greatly reduces the amount of image data required for learning. Then, we propose a MedQFormer, which contains learnable queries to allow aligning visual features with textural ones desired by a language model. Lastly, we choose BioMedLM~\cite{venigalla2022biomedlm} as our basic language model and fine-tune it using the LoRA~\cite{hu2021lora} technique. Our CAD model MedBLIP is lightweight and trainable on a single NVIDIA RTX 3090 GPU.

%learning multi-model representations from paired images and texts for CAD. Lastly, we design a set of prompts, which augments the learning of a large language model to find stronger connections between images and texts. 

To train and evaluate the effectiveness of our proposed MedBLIP model, we collect more than 30,000 medical image volumes from five public AD datasets, including ADNI~\cite{petersen2010alzheimer}, NACC~\cite{beekly2007national}, OASIS~\cite{marcus2007open},
AIBL~\cite{ellis2009australian}, and MIRIAD~\cite{malone2013miriad}. 
%For each image, we generate its corresponding textural description, including the volumes of different brain structures, age, gender, scores, etc. 
After pre-training on most of the images from ADNI, NACC, and OASIS datasets, we evaluate our MedBLIP on two tasks: (1) zero-shot classification, which directly applies pre-trained MedBLIP to classify unseen subjects from AIBL and MIRIAD datasets into three classes, i.e., normal controls (NC), mild cognitive impairment (MCI), and AD; and (2) zero-shot medical visual question answering (VQA), which generates an initial diagnosis for an unseen AIBL or MIRIAD subject based on input images and text descriptions and also provides some reasons for making such decision.

Overall, our contributions of this paper are summarized below:
\begin{itemize}
    \item We propose a lightweight CAD system MedBLIP, which is pre-trained on electronic health records in the form of images and texts, performs zero-shot classification, and makes medical VQA. The architecture of our CAD system is general and has the potential to incorporate more modalities and extend to other diseases beyond AD. 

    \item We propose a MedQFormer module, which extracts 3D medical image features and aligns them with textural features to be fused into a language model (LM). This module provides a way to align different types of medical data into the common space of LM, which is generic and could be used in other medical applications. 

    \item To our best knowledge, we have collected the largest public dataset for studying AD. On this dataset, our MedBLIP achieves the SOTA performance on separating AD and MCI subjects from healthy controls. Besides, we directly work on raw images without any preprocessing, which makes our system easy to use in practice.

    %3D medical vision model can be bootstrapped by a small number of parameter updates to an already-trained natural 2D natural vision model, with our proposed query mechanism called MedQformer;
    
    %\item Our approach exploits this domain knowledge by learning to incorporate images and correlate them to reports, leading to pre-trained models that can generalise to a wider range of embedding representation;

   % \item Inspired by BLIP-2~\cite{li2023blip}, powered by LLMs (e.g. OPT, FlanT5), MedBLIP can be prompted to perform zero-shot image-to-text generation that follows natural language instructions, which enables emerging capabilities such as visual knowledge reasoning with its generally knowledge in medical related fields;
    
    %\item Zero shot clasification, Finetune classification, Image-text retrival. SOTA

\end{itemize}

\section{Related Works}

\textbf{Vision Language Pre-Training.}
Data collected from different modalities typically provide different views about the data, which often complement each other and provide more complete information to facilitate a holistic understanding of the data. Vision-language pre-training (VLP) aims to learn multimodal foundation models, showing improved performance on various vision-and-language tasks~\cite{radford2021learning}. Roughly, we can divide current VLP models into two categories when fusing multi-modal inputs: light fusion and heavy fusion.

%There are currently two main approaches to multi-modal tasks: light fusion and heavy fusion. 
The approaches in the light fusion category focus on multi-modal alignment, which facilitates text matching, retrieval, and other downstream tasks, with representative methods like CLIP~\cite{radford2021learning} and ALIGN~\cite{jia2021scaling}. These methods directly align image representations with the corresponding text representations using a contrastive loss. DeCLIP~\cite{li2021supervision} exploits inter/intra-modality supervision to train a CLIP-like model with fewer data. On the other hand, the heavy fusion category focuses on incorporating multi-modal information with an attention mechanism to perform additional tasks. For instance, ALBEF~\cite{li2021align} proposes a contrastive alignment, which is followed by deeper fusion with a multi-modal encoder. Methods such as BLIP~\cite{li2022blip}, MoCo~\cite{he2020momentum}, CoCa~\cite{yu2022coca} incorporate a decoder and add image-to-text generation as an auxiliary task. Heavy fusion can interpret VQA, captions, and other downstream tasks that require more information for fusion and understanding.

%while the light fusion algorithms probably have difficulty to handle the same cases. %However, this approach is not as efficient in retrieval.

Medical image-text representation learning has been investigated based on contrastive learning as well. 
CheXzero~\cite{tiu2022expert} directly applies the CLIP on large-scale chest X-ray datasets to enable zero-shot classification of unseen findings in images.
MedCLIP~\cite{wang2022medclip} decouples paired images and texts and uses soft targets of semantic similarities to learn from unpaired medical images and texts.
BioViL-T~\cite{bannur2023learning} proposes a novel multi-image encoder to augment the current image representation with information from previous images. Most existing medical VLP are designed based on 2D images, since compared to 3D image volumes, 2D slices are sufficient to form a large-scale dataset for learning. However, in this paper, we aim to develop a medical VLP based on 3D image volumes with relatively few parameters and limited data size, i.e., a lightweight medical VLP for learning a 3D medical image and text representation. 

\textbf{LLMs for Multi-Modal Understanding.} Recently, using large language models (LLMs) as decoders in vision-language tasks has gained significant attention. This approach takes advantage of cross-modal transfer, which allows sharing knowledge between language and multi-modal domains.
VisualGPT~\cite{chen2022visualgpt} and Frozen~\cite{tsimpoukelli2021multimodal} have demonstrated the advantage of employing a pre-trained language model as a vision-language model decoder. 
Flamingo~\cite{alayrac2022flamingo} freezes a pre-trained vision encoder and language model and then fuses vision and language modalities with gated cross-attention. 
BLIP-2~\cite{li2023blip} designs a Q-Former to align the visual features from the frozen visual encoder with large language models, like FLAN-T5~\cite{chung2022scaling} and OPT~\cite{zhang2022opt}. FROMAGe~\cite{koh2023grounding} freezes large language models and visual encoders, and fine-tunes linear mapping layers to achieve cross-modality interactions. This method shows strong zero-shot performances on contextual image retrieval and multi-modal dialogue tasks. Built upon PaLM~\cite{chowdhery2022palm}, PaLM-E~\cite{driess2023palm} employs features from sensor modalities and integrates real-world continuous sensor modalities into an LLM, thereby establishing a connection between real-world perceptions and human languages. 
GPT-4~\cite{openai2023gpt4} presents powerful visual understanding and reasoning abilities after pre-training on a vast collection of image-text data. 

Most recently, several domain-specific multi-modal LLMs have been developed. ChatCAD~\cite{wang2023chatcad} combines visual and linguistic information processed by various networks as inputs of large language models to develop a medical-image CAD model, which provides a condensed report and offers interactive explanations and medical recommendations. Open-ended MedVQA~\cite{van2023open} employs a multi-layer perceptron (MLP) network that maps the extracted visual features from a frozen vision encoder to a set of learnable tokens, which develops an open-ended VQA for diagnoses and treatment decisions. 
% chatcad 是一个很大的模型；medvqa轻量但不align,他们也都是二维的
Differently, our MedBLIP explores a lightweight framework that works on 3D medical scans and aligns different types of medical data for CAD.

\section{MedBLIP}

\subsection{Problem Formulation}
We design a CAD system in the form of dialogue, with application to automatic AD diagnosis. Given inputs of a brain image scan $I$ collected from a subject and a textual description $T$ generated from this subject's EHRs in natural language, for a question asked in natural language $Q$, our CAD aims to sequentially generate an answer $A = \{A_0, A_1, ..., A_N \}$ composed of $N$ tokens, by conditioning on all inputs $\{I, T, Q\}$. To achieve this goal, we build a CAD model based on a large language model and find its optimal parameters $\theta^{*}$ by maximizing the conditional log-likelihood below:
%From a model definition perspective, we aim to find the optimal parameters $\theta^{*}$ for a model by maximizing the conditional log-likelihood as follows:
\begin{equation}
    \theta^{*}={\arg \max }_{\theta} \sum_{i=1}^{N} \log p_{\theta}\left(\mathbf{A}_{i} \mid \mathbf{I}, \mathbf{T}, \mathbf{Q}, \mathbf{A}_{i-1}\right). 
\end{equation}

\subsection{Network Framework}
Our CAD model is designed as an encoder-decoder architecture, with a two-stream encoder and a language model (LM) as a decoder, as illustrated in Fig.~\ref{fig:cad}. Specifically, the two-stream encoder takes inputs from two modalities, namely a vision sub-encoder for the image $I$ and the text sub-encoder for the textual description $T$ and the question $Q$. The language model is defined as a causal language transformer, which generates the answer $A$ in an auto-regressive manner.

\textbf{Vision Encoding Stream.} 
To encode a brain image volume and fully leverage an existing large model for reducing data requirements, we employ a pre-trained 2D vision encoder to extract its 3D visual features. % $\{x_1,x_2,\cdots,x_{l_x}\}$. 
To make this work, we need to address two problems: (1) bridging the domain and dimension gaps between a 2D vision encoder and a 3D medical scan, and (2) aligning image features with textural ones, which allows mapping all inputs into the latent space of the LM decoder for learning multi-modal representations. Inspired by~\cite{li2023blip}, we propose a query network based on a transformer encoder, which maps the visual features into a visual prefix $H_v = \{v_1,v_2,\cdots,v_{\ell_{v}}\} \in \mathbb{R}^{\ell_{v} \times e}$ for the language model, where $\ell_{v}$ is the length of the vision embedding sequence and $e$ is the embedding size. Also, we have a lightweight projection, which is learnable and adapts 3D image volumes to inputs of a pre-trained image encoder. This medical query transformer (MedQFormer) tackles the above two problems and will be discussed in detail in Sect.~\ref{sec:medqformer}.

%To use these features as input to the LM decoder, they should be mapped into the latent space of the language decoder. 
%Following~\cite{li2023blip}, we design a query network, implemented as a transformer encoder. This network maps the visual features into a visual prefix $\{v_1,v_2,\cdots,v_x\} \in \mathbb{R}^{\ell_{x} \times e}$ for the language model, where $e$ is the embedding size. 
%The specific implementation of query network will be illustrated in Section~\ref{sec:medqformer}.

\textbf{Language Encoding Stream.} Regarding the textural description of subjects' EHRs except for image scans and the asked questions, we first utilize a standard tokenization process as in~\cite{jain2022hugging} to obtain a sequence of tokens, i.e., 
%firstly we utilize a standard tokenization process to obtain a sequence of tokens, for 
the textual description $\mathbf{T}= \{t_1,t_2,\cdots,t_{\ell_t}\} \in \mathbb{R}^{\ell_{t} \times e}$, the question $\mathbf{Q}= \{q_1,q_2,\cdots,q_{\ell_q}\} \in \mathbb{R}^{\ell_{q} \times e}$, and the answer $\mathbf{A}= \{a_1,a_2,\cdots,a_{\ell_a}\} \in \mathbb{R}^{\ell_{a} \times e}$, where $\ell_{t}$, $\ell_{q}$, $\ell_{a}$ indicate the length of the embedding sequence of the text, question, and answer, respectively.
% and  where
% \fixme{$\ell_{q}$ is the length of question embedding sequence}. 
% and  where 
% \fixme{$\ell_{a}$ is the length of answer embedding sequence}.
% why do we need to encode answers?}. 
% 因为要计算LLM的loss,answer要编码和模型输出算loss的
These tokens are embedded later using the embedding function provided in a pre-trained language model. 

%This is followed by embedding the tokens using the embedding function of a pre-trained language model.

\textbf{Prompt Structure.}
To create a structured prompt, following current VQA methods used in language models~\cite{li2023blip,van2023open}, we prepend the question and answer %要有‘Answer： ’这个prompt才容易生成回答 
tokens with tokenized descriptive strings, namely in the form of \textbf{question:} and \textbf{answer:}. We choose to place the embeddings of the image and text description before the question tokens. As a result, we have the following prompt template: 
\begin{equation}
    \begin{aligned}
    p =& [ v_1,v_2,\cdots,v_{\ell_x}, t_1,t_2,\cdots,t_{\ell_t}, \mathbf{Question:} What\ \\
    & will\ this\ subject\ be\ diagnosed\ with? \textbf{Answer:} ],
    \end{aligned}
\end{equation}
which is fed as input to the language model below.

\textbf{Language Model.} 
Following standard language modeling systems~\cite{venigalla2022biomedlm}, we treat VQA as a process of a conditional generation of text, and we optimize the standard maximum likelihood objective during training. The language model receives the prompt sequence as input and outputs the answer $A$, token by token. Specifically, at each time step $i$, the outputs of the model are the logits, which parameterize a categorical distribution $p_{\theta}(A)$ over the vocabulary tokens. This distribution is represented as follows:
\begin{equation}
    \log p_{\theta}(\mathbf{A})=\sum_{l_{a}} \log p_{\theta}\left(a_{i} \mid v_{1}, \ldots v_{\ell_v}, t_{1}, \ldots t_{\ell_{t}}, q_{1}, \ldots q_{\ell_{q}}, a_{1}, \ldots a_{i-1}\right).
\end{equation}
%\fixme{where $l_a$ is the length of answer embedding sequence.}
The parameters of the language model are initialized from a pre-trained model, which has been previously pre-trained on huge web-collected datasets~\cite{dodge2021documenting,gao2020pile}.

\subsection{MedQFormer}
\label{sec:medqformer}

\begin{figure*}
  \centering
  \includegraphics[width=0.86\linewidth]{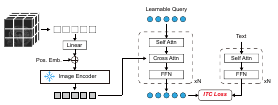}
  \caption{Illustration of our proposed MedQformer that aligns 3D visual and textural features for learning in the unified latent space of language model.}
  \label{fig:medqformer}
\end{figure*}

To bridge the gap between 3D medical images and 2D vision encoders pre-trained on natural images, inspired by BLIP-2~\cite{li2023blip}, we employ a query encoder to extract and align vision features.

\textbf{Image Feature Extraction.}
We first divide the input image $I$ into a set of 3D sub-volumes $\{Iv_i\}_{i=1}^{N_v}$, followed by a linear projection $f_{\varphi_1}$ that projects 3D cubes into 1D image embeddings $\{E_i = f_{\varphi_1}(Iv_i)\}_{i=1}^{N_v}$. With the addition of learnable position embeddings $f_{\varphi_2}$, the image embeddings can be received as inputs of a standard pre-trained vision encoder to extract desired image features. Although the pre-trained vision encoder $f_{\phi}$ has fixed parameters $\phi$, we have learnable linear projection and position embedding to transfer a 2D vision encoder to a 3D medical domain. Hence, we have a medical vision encoder with learnable parameters $\varphi_1$ and $\varphi_2$, which maps a volumetric image $I$ into $N_v$ visual features $f_1,\cdots,f_{N_v} = \{f_{\phi}(f_{\varphi_1}(Iv_i), f_{\varphi_2}(Iv_i))\}_{i=1}^{N_v}$. As a result, we obtain the final image embeddings $IE = (f_i, \cdots, f_{N_v})$ for each input image volume $I$.

%$f_1,\cdots,f_{N_v} = f_{\varphi_1, \varphi_2, \phi}(\mathbf{\fixme{E? why not I?}})= \{f_{\phi}(f_{\varphi_1}(Iv_i), f_{\varphi_2}(Iv_i))\}_{i=1}^{N_v}$.

\textbf{Query Encoder.}
To map the visual features $\{f_i\}_{i=1}^{N_v}$ into the common language space, we use a set of $L$ learnable queries $qry_i \in \mathbb{R}^{d_e}$, where $d_e$ is the dimension of query embeddings. These queries have a transformer structure that interacts with the image encoder for adjusting visual feature extraction and a text transformer as a textural feature extractor. As shown in Fig.~\ref{fig:medqformer}, these learnable queries interact with each other through self-attention layers, then interact with image features through cross-attention layers. As a result, we obtain a visual prefix $H_v$ that is aligned with textural features and can be taken by a language model.

\subsection{Training MedBLIP}

\textbf{Learnable Parameters.}
Standard fine-tuning of a language model could hurt its generalization capability, especially if a dataset used for fine-tuning has a small size and is domain-specific as in our case. Therefore, we consider two parameter-efficient strategies that adapt the attention blocks of language models:
\begin{itemize}
    \item \textbf{Frozen LM.} 
The parameters of the language model are kept entirely frozen during training. In this setting, only the 3D vision query network is updated through backpropagation.
    \item \textbf{Low-Rank Adaptation (LoRA).} 
We add learnable weight matrices to the query $Q_w$ and value $V_w$ of the attention blocks in each layer of the frozen language model as $W + \triangle W$ following~\cite{hu2021lora}. In this setting, the 3D vision query network function is trained together with the learnable weight matrices.
\end{itemize}

\textbf{Objective Functions.}
We have loss functions for MedQformer and LM modules in our MedBLIP model. As discussed in Sect.~\ref{sec:medqformer}, MedQformer includes both a transformer image encoder $E_I$ and a transformer text encoder $E_T$. During training, we have a set of image-text pairs $(I, T)$ and a set of image and diagnosis Q\&A pairs (I, Q\&A). We use the image-text contrastive learning (ITC) loss in~\cite{radford2021learning} to align multi-modal representation, resulting in our feature alignment loss:
\begin{equation}
    \mathcal{L}_{FA}=contrastive\left(E_{I}(I), E_{T}(T)\right) + {contrastive}\left(E_{I}(I), E_{T}(Q\&A)\right).
    \label{eq:fa_loss}
\end{equation}
Similar to BLIP-2~\cite{li2023blip}, we select the one that has the highest similarity with text from multiple output query embeddings to compute the ITC Loss.
To supervise the LM component, we use cross entropy to compute language generation loss $\mathcal{L}_{{LG}}$. Hence, the final loss function is defined as:
\begin{equation}
    \mathcal{L}_{total}=\mathcal{L}_{FA}+\lambda_{LG} \mathcal{L}_{LG},
\end{equation}
where $\lambda_{LG}$ is a hyperparameter to balance these two terms. 
%the feature alignment loss $\mathcal{L}_{{FA}}$ and the language generation loss $\mathcal{L}_{{LG}}$. 

\section{Experiments}

\begin{table*}
\scriptsize
\setlength{\tabcolsep}{1.6pt}
  \caption{Demographic statistics of used AD datasets. \small{F: female, M: male, Educ: Education level, SES: Socio-Economic Status, MMSE: Mini-Mental State Examination, CDR: Clinical Dementia Rate, E/L/S/PMCI: early, late, stable, and progressive MCI, IMCI: Impaired not MCI, and DEM: demented. \# indicates the number.}}
  \label{tab:data}
  \begin{tabular}{c|ccccccccc}
    \toprule
    \multirow{2}*{\textbf{Datasets}} & \multirow{2}*{\textbf{\#Images}} & \multicolumn{8}{c}{\textbf{Texts}} \\
    \cline{3-10}
     & & \#F/\#M & Age(\#) & Educ(\#) & SES(\#) & MMSE(\#) & CDR(\#) & Logical Memory(\#) & Diagnosis(\#) \\
    \midrule
    ADNI & 10387 & 4710/5677 & 45-95(10386) & 9860 & - & 9385 & 9401 & 7189 & NC,MCI,E/L/S/PMCI,AD (10387) \\
    NACC & 15354 & 9058/6296 & 19-102(15354) & 15329 & - & 7867 & 15354 & 7654 & NC, IMCI, MCI, DEM (14277) \\
    OASIS & 3020 & 1798/1222 & 18-98(3020) & 2300 & 2153 & 2293 & 2300 & - & DEM, Non-DEM (336) \\
    AIBL  & 1002 & 471/531 & 42-96(1002) & - & - & 1002 & 1002 & 1002 & NC, MCI, AD (997) \\
    MIRIAD & 708 & 393/315  & 55-87(708) & - & - & 268 & 46 & - & NC, AD(708) \\
  \bottomrule
\end{tabular}
\end{table*}

\subsection{Datasets and Experimental Settings}

We collect more than 30,000 image volumes from five public datasets for studying AD/Dementia and evaluate our CAD system MedBLIP on separating subjects with AD or mild cognitive impairment (MCI) from normal controls (NC). Table~\ref{tab:data} reports the demographic statistics of these five datasets. 

%are the current most suitable Alzheimer's Disease datasets given their large variety in EHRs. See the datatset details in Table~\ref{tab:data}.

% 标红的地方回来再改
\textbf{ADNI~\cite{petersen2010alzheimer}}. 
% The Alzheimer’s Disease Neuroimaging Initiative (ADNI) is a longitudinal multicenter study, which is designed to develop clinical, imaging, genetic, and biochemical biomarkers for the early detection and tracking of Alzheimer’s disease. 
This dataset has 10,387 volumetric T1 MRI scans that went through a series of pre-processing steps, including denoising, bias field correction, skull stripping, and affine registration to the SRI24 atlas, with an image size of $138 \times 176 \times 138$. For testing, we subject-wisely sample a subset of 200 images in each class (i.e., NC, MCI, AD), which is named ADNI-3x200.
% \fixme{how many subjects are included to form a set of 200 images?}

%We use the training split of this dataset for pre-training. For evaluation, we sample subject-wisely a multi-class classification dataset from the testing split, namely ADNI-3x200. This multi-class classification dataset has 200 exclusively positive images for the three ADNI competition tasks: Normal cognition, MCI, and dementia.

\textbf{NACC~\cite{beekly2007national}}. 
% The National Alzheimer's Coordinating Center (NACC) maintains a large relational database of standardized clinical and neuropathological research data and 
This dataset has a large amount of raw volumetric T1 MRI scans with a variety of resolutions. We select those MRIs having 100$\sim$256 slices in all three dimensions, resulting in 15,354 images. Unlike the ADNI dataset, we directly use the raw data; but similarly, we sample subject-wisely a NACC-3x200 dataset for testing. 

%We use the training split of this dataset for pre-training. For evaluation, we also sample subject-wisely a NACC-3x200 dataset for the same three tasks above.

\textbf{OASIS~\cite{marcus2007open}}. 
% The Open Access Series of Imaging Studies (OASIS) is a project that aims at making neuroimaging datasets of the brain freely available to the scientific community. 
We collect 3020 volumetric T1 MRIs from OASIS 1\&2. These scans went through pre-processing with denoising and skull stripping and have a size of $256 \times 256 \times 256$. Since OASIS 1 only releases some clinical reports but with no diagnoses (e.g. NC, MCI or dementia), we use all images from OASIS 1 for pre-training. For testing, we sample subject-wisely an OASIS-2x200 subset from OASIS 2 to separate demented and non-demented subjects. 

%for OASIS1 and $256 \times 256 \times 256$ for OASIS2. They have . 

\textbf{AIBL~\cite{ellis2009australian}}. 
% The Australian Imaging, Biomarker \& Lifestyle Flagship Study of Ageing (AIBL) is a study to discover that which biomarkers, cognitive characteristics, and health and lifestyle factors determine subsequent development of symptomatic AD. 
This dataset has 1002 volumetric T1 MRI scans with sizes of $160 \times 240 \times 256$, which are collected from demented, MCI, or healthy subjects. We do not use this data for training; for testing, we sample a balanced subset with 200 images each for NC, MCI, and dementia classes.
%We collect raw volumetric T1 MRI scans from official website and sample a balanced subset(i.e., 1:1:1), resulting $3\times200$ images for test.

% We collect 1002 raw volumetric T1 MRI scans in total. These images have a resolution of $160 \times 240 \times 256$. We use all images from this dataset for test.

\textbf{MIRIAD~\cite{malone2013miriad}}. 
% The Minimal Interval Resonance Imaging in Alzheimer's Disease (MIRIAD) dataset is a series of longitudinal 
% volumetric T1 MRI scans of 46 mild-moderate Alzheimer's subjects and 23 controls. 
% All subjects were requested to attend five to ten imaging visits from baseline to 2 years. 
We collect 708 raw volumetric T1 MRI scans, which have an image size of $124 \times 256 \times 256$. This is a binary classification dataset with two labels, i.e., demented and not-demented subjects. We sample a balanced subset with a 1:1 positive and negative ratio, resulting in $2\times200$ images for testing. No images are used for training to perform zero-shot experiments.

As a result, we have most images from ADNI, NACC, and OASIS datasets for pretraining and save images from AIBL and MIRIAD datasets for zero-shot testing. In total, we held 1000 subjects with 2600 samples out for evaluation. To simplify the preprocessing step, all images are first padded to a cube and then scaled to a unified size of $224 \times 224 \times 224$ as inputs. 

\textbf{Implementation Details.}
For the frozen image encoder, we choose state-of-the-art pre-trained ViT-G/14 from EVA-CLIP~\cite{fang2022eva}, which is demonstrated to be effective in BLIP-2~\cite{li2023blip}. For the input image with a size of $ 224 \times 224 \times 224 $, the patch size and the stride are both set as 32, resulting in image features with the size of $344\times1408$. For the MedQformer, we use 32 learnable queries, where each query has a dimension of 768 and the hidden layers $N$ is set to 12. Regarding language models, we have three options, i.e., FLAN-T5~\cite{chung2022scaling}, BioGPT~\cite{luo2022biogpt}, and BioMedLM~\cite{venigalla2022biomedlm}. 
FLAN-T5 is an instruction-trained model with 3B parameters trained on C4 WebText~\cite{dodge2021documenting}. BioGPT and BioMedLM  are both GPT models relying on GPT-2 architecture, pre-trained on PubMed and biomedical data from the Pile~\cite{gao2020pile}, with a size of 1.5B and 2.7B parameters, respectively. All our models are able to fine-tune on a single NVIDIA RTX 3090 GPU. We use the AdamW optimizer with a learning rate of 5e-3. The hyperparameter $\lambda_{LG}$ is set to 1.

% We compare MedBLIP with four SOTA language-models, i.e., pre-trained on a general text corpus, such as GPT-3 and FLAN-T5, and medically-trained models, such as BioGPT and BioMedLM. 

% We also provide ablation study with ResNet-50~\cite{} as the vision encoder, which is in-line with previous works~\cite{}.

% \textbf{CLIP}~\cite{radford2021learning} A vision-text contrastive learning framework pre-trained on a dataset of 400M image-texts pairs collected from the internet.

% \textbf{MedCLIP}~\cite{wang2022medclip} A decoupled medical image-text contrastive learning framework pre-trained on a 570k medical image-texts pairs collected from large datasets of chest X-rays: CheXpert and MIMIC-CXR.

% \textbf{BLIP2}~\cite{li2023blip} A vision-language pre-training method which leverages frozen pretrained image encoders and LLMs, with capabilities in zero-shot instructed image-to-text generation.

% In this paper, we adopt the pre-trained image encoder ViT-G/14 used in CLIP~\cite{radford2021learning} as our frozen image encoder in MedBLIP. For the frozen language model, we apply the instruction-trained FlanT5~\cite{chung2022scaling} for encoder-decoder-based LLMs.

\begin{table*}[t]
\centering
\scriptsize
  \caption{Experimental results of our MedBLIP on five datasets, including zero-shot CAD on the last two datasets. The classification performance is measured in the mean accuracy (ACC) with five runs. The best scores are in bold.}
  \label{tab:exp1}
  \begin{tabular}{cc|cc|ccccc}
    \toprule
    \multicolumn{2}{c|}{\multirow{2}{*}{Methods}} & LM & {\#Learnable} & ADNI & NACC & OASIS & \multirow{2}{*}{AIBL} & \multirow{2}{*}{MIRIAD} \\
     & & size & params & -3x200 & -3x200 & -2x200 &  &  \\
    \midrule
    FLAN-T5\tiny{~\cite{chung2022scaling}} & Text only & \multirow{3}{*}{3.4B} & - & 37.0\% & 39.5\% & 46.7\% & 33.3\% & 60.0\% \\
    \multirow{2}{*}{Ours w/ T5} & Frozen &  & 151M & 50.5\% & 69.2\% & 61.3\% & 54.7\% & 64.0\% \\
    & LoRA & & 156M & 64.0\% & 77.3\% & 75.8\% & 59.2\% & 66.8\% \\
    \midrule
    BioGPT\tiny{~\cite{luo2022biogpt}} & Text only & \multirow{3}{*}{1.5B} & - & 25.7\% & 21.7\% & 28.3\% & 26.7\% & 50.0\% \\
    \multirow{2}{*}{Ours w/ BioGPT} & Frozen &  & 151M & 56.3\% & 66.5\% & 66.0\% & 60.7\% & 55.2\% \\
    & LoRA & & 156M & 62.2\% & 72.3\% & 71.7\% & 62.4\% & 59.7\% \\
    \midrule
    BioMedLM\tiny{~\cite{venigalla2022biomedlm}} & Text only & \multirow{3}{*}{2.7B} & - & 62.5\% & 63.5\% & 61.8\% & 65.7\% & 46.3\% \\
    \multirow{2}{*}{Ours w/ BioMedLM} & Frozen &  & 151M & 71.2\% & 82.0\% & 79.8\% & 77.8\% & 66.1\% \\
    & LoRA & & 154M & \cellcolor{lightgray}\textbf{78.7\%} & \cellcolor{lightgray}\textbf{83.3\%} & \cellcolor{lightgray}\textbf{85.3\%} & \cellcolor{lightgray}\textbf{80.8\%} & \cellcolor{lightgray}\textbf{71.0\%} \\
  \bottomrule
\end{tabular}
\end{table*}

\begin{figure*}[t]
  \centering
  \includegraphics[width=\linewidth]{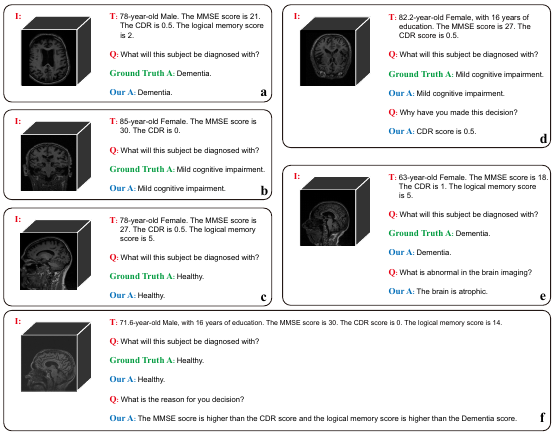}
  \caption{Samples of zero-shot results on the AIBL dataset, which are generated by our MedBLIP built on BioMedLM with LoRA fine-tuning. 
  % \fixme{ours A --> Our A:, What made you make this decision? --> Why did you make this decision? or Why have you made this decision?}
  }
  \label{fig:vis}
\end{figure*}

\begin{table}[t]
\footnotesize
\centering
\caption{Comparison between a large vision encoder and our MedQFormer on ADNI.}
\label{tab:ab1}
  \begin{tabular}{c|c|c}
  \toprule
  Visual features & \#Params & Accuracy \\
  \midrule
  ViT-G~\cite{fang2022eva} & 1B & 72.2\% \\
  Our MedQFormer & 151M & 71.6\% \\
  \bottomrule
  \end{tabular}
\end{table}

\begin{table}[h]
\footnotesize
\centering
\caption{Comparison between different prompt structures.}
\label{tab:ab2}
  \begin{tabular}{c|ccccc}
  \toprule
  \multirow{2}{*}{Setting} & ADNI & NACC & OASIS & \multirow{2}{*}{AIBL} & \multirow{2}{*}{MIRIAD} \\
  & -3x200 & -3x200 & -2x200 &  & \\
  \midrule
  Regular (I\&T, Q, A) & 78.7\% & 83.3\% & 85.3\% & 80.8\% & 71.0\% \\
 Alternative (Q, I\&T, A) & 79.3\%(\textcolor{red}{+0.6}) & 82.8\%(\textcolor{blue}{-0.5}) & 82.5\%(\textcolor{blue}{-1.8}) & 82.8\%(\textcolor{red}{+2.0}) & 70.8\%(\textcolor{red}{+0.2}) \\
  \bottomrule
  \end{tabular}
\end{table}

\begin{table}%[h]
\scriptsize
\centering
\caption{Ablation study on loss functions. }
\label{tab:ab3}
  \begin{tabular}{c|ccccc}
  \toprule
  \multirow{2}{*}{Loss Function} & ADNI & NACC & OASIS & \multirow{2}{*}{AIBL} & \multirow{2}{*}{MIRIAD} \\
  & -3x200 & -3x200 & -2x200 &  & \\
  \midrule
  $contrastive(I,T)$ & 71.7\% & 80.5\% & 82.5\% & 74.7\% & 66.8\% \\
 $contrastive(I,T) + contrastive(I,Q\&A)$ & 78.7\%(\textcolor{red}{+7.0}) & 83.3\%(\textcolor{red}{+2.8}) & 85.3\%(\textcolor{red}{+2.8}) & 80.8\%(\textcolor{red}{+6.1}) & 71.0\%(\textcolor{red}{+4.2}) \\
  \bottomrule
  \end{tabular}
\end{table}

\subsection{Experimental Results}

% We conduct extensive experiments on Five MRI datasets to answer the following questions:

% \textbf{Q1.} Does the proposed pre-training method yield better zero-shot medical image classification performances?

% \textbf{Q2.} Does \textbf{MedBLIP} bring better performance and label efficiency for downstream classification tasks with fine-tuning?

% \textbf{Q4.} Are the learned embeddings good at Zero-shot Medical VQA tasks?

% \textbf{Q5.} How do the learned embeddings look like?

% MedBLIP enables the LLM to understand images while preserving its capability in following text prompts, which allows us to control image-to-text generation with instructions, and apply to the downstream work. We simply append the text prompt after the visual prompt as input to the LLM. We use "Question: \{ \} Answer:" as the prompt. 
% To perform quantitative evaluation on the zero-shot classification task and align better with human annotation, we constructed the problem of this choice property, e.g., "Is the patient Healthy, MCI, or Dementia?" Then the learned multi-modal representations are used to support zero-shot prediction.

% We conduct zero-shot classification evaluation on 5 datasets: ADNI-3x200, NACC-3x200, OASIS-2x200, AIBL, and MIRIAD. As shown in Table~\ref{tab:exp1}, MedBLIP achieves state-of-the-art result on the five datasets. 

% Figure~\ref{fig:examples} shows examples to demonstrate a wide range of zero-shot Medical VQA. 
\textbf{Zero-shot Medical CAD.}
Table~\ref{tab:exp1} reports the evaluation of our MedBLIP
using different language models and settings. The language models, i.e., FLAN-T5, BioGPT, and BioMedLM, show their capability of performing monomodal medical CAD, i.e., using text descriptions only, to some extent. Among these three language models, BioMedLM performs the best, showing that it captures some dependencies between prompts and inherent knowledge when generating answers. By adding the visual modality even without fine-tuning, the performance of our model on all datasets has improved significantly. The accuracy improvement varies within [4.0\%, 44.8\%], and the BioGPT benefits the most from the visual input. This result indicates the necessity of using image scans in diagnosis. Using the fine-tuning technique LoRA, our performance is further improved, with at least 1.3\% and at most 14.5\% improvement in accuracy. Overall, our MedBLIP built upon BioMedLM and LoRA fine-tuning shows the best performance on all datasets. 

% The evaluation of our method across various language models and fine-tuning settings in Table~\ref{tab:exp1} shows that language models can perform unimodal medical CAD due to the capability of pre-trained language models to capture dependencies between prompt and inherent knowledge when generating answers. With the addition of visual modality, the performance has improved. Additionally, LoRA improves the performance of the model which directly adapts the Q and V weight matrices of the attention blocks.
Figure~\ref{fig:vis} (a-c) visualizes the zero-shot CAD process on unseen subjects sampled from the AIBL dataset. Take Fig.~\ref{fig:vis}(b) for example, although the text description of this subject shows no significant difference from those of healthy subjects, in brain scans the hippocampus and ventricle show the presence of abnormal atrophy. Our MedBLIP provides the correct diagnosis of MCI.

% \begin{table*}[t]
% \centering
% \scriptsize
%   \caption{Results of zero-shot CAD tasks on five datasets. Performance across different language models and fine-tuning strategies, measured in the mean accuracy (ACC) in five runs. Best scores across a dataset are in bold.}
%   \label{tab:exp1}
%   \begin{tabular}{cc|cc|ccccc}
%     \toprule
%     \multicolumn{2}{c|}{Methods} & LM size & Params & ADNI-3x200 & NACC-3x200 & OASIS-2x200 & AIBL & MIRIAD \\
%     \midrule
%     \multicolumn{2}{c|}{BioGPT~\cite{}} & 1.5B & - & 25.7\% & 21.7\% & 28.3\% & 26.7\% & 50.0\% \\
%     \multicolumn{2}{c|}{BioMedLM~\cite{}} & 2.7B & - & 62.5\% & 63.5\% & 61.8\% & 65.7\% & 46.3\% \\
%     \multicolumn{2}{c|}{FLAN-T5~\cite{}} & 3.4B & - & 37.0\% & 39.5\% & 46.7\% & 33.3\% & 60.0\% \\
%     \midrule
%     \multirow{2}{*}{Ours w/ BioGPT} & Frozen & \multirow{2}{*}{1.5B} & & 56.3\% & 66.5\% & 66.0\% & 60.7\% & 55.2\% \\
%     & LoRA & & & 62.2\% & 72.3\% & 71.7\% & 62.4\% & 59.7\% \\
%     \midrule
%     \multirow{2}{*}{Ours w/ T5} & Frozen & \multirow{2}{*}{3.4B} & & 50.5\% & 69.2\% & 61.3\% & 54.7\% & 54\% \\
%     & LoRA & & & 64.0\% & 77.3\% & 75.8\% & 59.2\% & 66.8\% \\
%     \midrule
%     \multirow{2}{*}{Ours w/ BioMedLM} & Frozen & \multirow{2}{*}{2.7B} &  & 71.2\% & 82.0\% & 79.8\% & 77.8\% & 66.1\% \\
%     & LoRA & & & \cellcolor{lightgray}78.6\% & \cellcolor{lightgray}83.3\% & \cellcolor{lightgray}85.4\% & \cellcolor{lightgray}80.6\% & \cellcolor{lightgray}71.0\% \\
%   \bottomrule
% \end{tabular}
% \end{table*}

\textbf{Zero-shot Medical VQA.}
% 定性分析
Figure ~\ref{fig:vis} (d-f) shows the zero-shot Medical Visual question answering(VQA) ability of our MedBLIP. Since our approach is generative, after a simple classification-based question, MedBLIP provides a natural way of performing VAQ and presents the chain of thoughts.  
%this simple yet effective method provides a more natural way of doing VQA about the chain of thoughts.
% 生成模型的还有一个准确推理路径的问题
MedBLIP may also generate unsatisfactory answers to users' questions due to various reasons, including inaccurate medical knowledge from the LLM, activating the incorrect reasoning path, or not having up-to-date information about new image content.

%, As shown in Figure~\ref{fig:vis}(d-f).

\textbf{Ablation Study.} We perform ablation studies from three aspects to answer the following three questions: (1) Why use a 2D pre-trained vision encoder instead of a trainable large vision encoder? (2) Will a prompt structure make a difference in the final CAD result? and (3) Why need the ITC loss between the image and diagnosis Q\&A?  

% We ablate several design choices on MedBLIP in Table~\ref{}. 
% (1). Visual features.
% (2). Pretraining.
% (3). Model size.
\textbf{\it (1) Benefit of using a frozen 2D pre-trained vision encoder.} To demonstrate the effectiveness of our lightweight image encoder based on the 2D pre-trained model, we take the query output embedding from MedQformer and compare it with features extracted from trainable ViT-G~\cite{fang2022eva} on ADNI. We add a linear classification head with the cross-entropy loss. Table~\ref{tab:ab1} reports that MedQFormer achieves slightly reduced performances, i.e., 0.6\% lower than ViT-G in accuracy, but with much fewer parameters (only 15.1\% of ViT-G's). This lightweight module benefits downstream tasks and allows building our model on language models and training it on one GPU. We can also see that benefiting from this lightweight visual encoder, our MedBLIP outputs ViT-G by an improvement of 6.5\% in the classification accuracy on ADNI. 

%comparied with fine-tune all paramerters of network. However our MedQformer requires fewer parameters which is beneficial for downstream tasks.

\textbf{\it (2) Effect of using different prompt structures.} To answer the second question above, we investigate the order of three prompting components, i.e., image and text features, the question, and the answer, and its effect on our model's performance. We treat the one with the question in the middle as the regular prompt structure and compare it to the one starting with the question. Table~\ref{tab:ab2} shows that on some datasets our MedBLIP prefers the regular prompt, but this is not always the case. We conclude that the prompt strategy will not make a huge difference in the final performance of our model.   

%We also investigate the influence of the prompt structure on the overall performance, demonstrated in Table~\ref{tab:ab2}. We conclude that prompt strategy contributes relatively little to the final performance.

\textbf{\it (3) Necessity of using two ITC loss functions.} Besides the regular ITC loss between image and text pairs, we have another one between image and diagnosis Q\&A, as presented in Eq.~\ref{eq:fa_loss}. Table~\ref{tab:ab3} demonstrates that by adding the second ITC loss function, the classification accuracy improves on all datasets. This result is consistent with our motivation of adding the ITC loss between image and diagnosis Q\&A, since it enforces the learnable queries to extract image features related to CAD.

%In Table ~\ref{tab:ab3}, we show that the ITC loss between image and Diagnosis Q\&A is also beneficial for Zero-shot Medical CAD. This result supports our intuition in designing the representation learning objectives: the ITC loss enforces the queries to extract visual features most relevant to the diagnosis results, thus improving vision-language alignment.

\section{Discussion and Conclusion}

In this paper, we propose a novel CAD system MedBLIP that fuses medical multi-modal data, i.e., image and text, from EHRs and shows its capability of performing zero-shot classification and medical VQA. Our MedBLIP introduces MedQFormer, a lightweight trainable 3D vision encoder that acts as a bridge between 3D medical images and a large frozen 2D vision encoder and as a bridge between 3D medical images and language models. Moreover, MedBLIP operates with low computational costs, as it smartly combines large pre-trained vision and language models with no need of training them from scratch or a large dataset in a specific medical domain. Our experiments demonstrate the effectiveness of our approach by outperforming several baselines, which sheds new light on further exploring medical multi-modal CAD.

%  没有真的看见大脑，描述一个大脑，我们想的是对海马体/脑室有一个描述。原因是训练数据里缺少这样的信息，目前也没有开源的数据集描述一个3D MRI scan，这是我们想创建的
%  生成模型不能保证他的逻辑链路是对的，让他自己学，可能会错误归因
%  做大脑退行性病变，应该考虑longitudinal 信息，后面应该作图像序列的输入
%  输入长度的考虑，因为医疗数据的多模态信息是非常多的，目前对于长语料的支持还不够

% This work attempts to extract 3D medical domain knowledge with learning from frozen 2D pretrain vision encoder. Hence, it significantly expands the ability to pretrain with limited data in other domain. 

% It still encounters failure cases where open-ended question 

% limited data for describing the 3D brain imaging

\textbf{Limitations and Future Work.}
LLMs can perform in-context learning given domain-specific few-shot examples. However, in our experiments with MedBLIP, we do not observe an improved VQA performance when asking something about the input brain scan context, even though it made the correct diagnosis. We attribute the unsatisfactory VQA results to the lack of corresponding textural descriptions of brain scans in our dataset. Without a description of a 3D brain MRI scan, the LLMs have difficulty describing what they "observe" in this image, such as the shrunken hippocampus or the enlarged ventricles. Currently, no such dataset or model is available to provide an image caption/description for a brain scan. We will explore this direction in our future work. 

Besides, degenerative diseases like AD are often studied in the longitudinal setting since longitudinal atrophy has probably happened at an early stage of AD, making it easier to separate MCI subjects from normal controls. In future work, we will extend our model to take longitudinal inputs and further improve our classification accuracy. In addition, in our experiments, we only consider two modalities, i.e., MRIs and texts, other medical data sources, like positron emission tomography (PET) images, and audio, are also useful in diagnosing AD. Fortunately, the architecture of our MedBLIP is flexible enough to incorporate additional modalities, which is also left as our future work.

%in-context learning capability in our pretraining dataset, which only contains brain imaging-EHRs pairs. The LLMs cannot learn from the dscription of brain imaging. Specifically, there is no open source dataset to describe a 3D brain MRI scan, such as the location and shape of its hippocampus and ventricles. We aim to create a similar dataset in future work.

%What's more, MedBLIP could have unsatisfactory results due to various reasons including inaccurate knowledge from the LLM, activating the incorrect reasoning path (see Figure~\ref{fig:vis}). Finally, for degenerative diseases, it is necessary to develop model the ability to analysis the longitudinal input. 

\bibliographystyle{unsrtnat}
\bibliography{refs}

% \section{Supplementary Material}

% Authors may wish to optionally include extra information (complete proofs, additional experiments and plots) in the appendix. All such materials should be part of the supplemental material (submitted separately) and should NOT be included in the main submission.

\end{document}